\documentclass[runningheads]{llncs}

\usepackage{graphicx}
\usepackage{amsmath}
\usepackage{multirow}
\usepackage{enumerate}
\usepackage{microtype}
\usepackage[caption=false]{subfig}
\usepackage{caption}
\begin{document}

\title{Computer Vision based Animal Collision Avoidance Framework for Autonomous Vehicles}
\titlerunning{Computer Vision based Animal Collision Avoidance Framework for AVs}

\author{Savyasachi Gupta\orcidID{0000-0002-9080-2028} \and Dhananjai Chand\orcidID{0000-0003-2870-1514} \and Ilaiah Kavati \orcidID{0000-0002-2659-2329}}
\authorrunning{S. Gupta et al.}

\institute{National Institute of Technology Warangal, India - 506004
}

\maketitle

\begin{abstract}
	Animals have been a common sighting on roads in India which leads to several accidents between them and vehicles every year. This makes it vital to develop a support system for driverless vehicles that assists in preventing these forms of accidents. In this paper, we propose a neoteric framework for avoiding vehicle-to-animal collisions by developing an efficient approach for the detection of animals on highways using deep learning and computer vision techniques on dashcam video. Our approach leverages the Mask R-CNN model for detecting and identifying various commonly found animals. Then, we perform lane detection to deduce whether a detected animal is on the vehicle’s lane or not and track its location and direction of movement using a centroid based object tracking algorithm. This approach ensures that the framework is effective at determining whether an animal is obstructing the path or not of an autonomous vehicle in addition to predicting its movement and giving feedback accordingly. This system was tested under various lighting and weather conditions and was observed to perform relatively well, which leads the way for prominent driverless vehicle’s support systems for avoiding vehicular collisions with animals on Indian roads in real-time.

\end{abstract}
\section{Introduction}
\label{sec:intro}
Animal detection has been an area of interest for wildlife photographers for finding animals in wildlife. However, it has found recent use for the development of autonomous vehicles because it has been observed that in countries such as India, animals are seen to be roaming freely on the roads, which leads to unwanted accidents \cite{SG_14_1}. As a result, it becomes vital to have an effective method for detecting animals on the roads.

\par Additionally, such a method can be used for the development of a support system that can help reduce the number of accidents caused due to vehicle-to-animal collision in autonomous vehicles. This support system may even perform better than the standard human reaction time to stop the manually controlled vehicle in the presence of an obstructing animal on the road and thus make riding an autonomous vehicle safer than manually driven vehicles \cite{SG_14_2}.

\par Currently, some existing methods provide animal detection already. For instance, Sharma and Shah \cite{SG_23_1} use Cascade Classifiers and Histograms of Oriented Gradients (HOG) for animal detection. However, this approach is limited by a few factors. Firstly, it only provides an accuracy of 82.5\% and secondly, it is limited to just cow detection, which is one of the many animals found on Indian roads. Burghardt and Calic \cite{SG_23_2} have introduced a methodology that depends on animals taking a pose and facing the camera for detection (i.e. similar to face detection for humans). The drawback of this approach is that animals are detected using facial recognition which may not always be the case by the video capturing device as the animals may not always be facing directly towards it. 

\par Furthermore, Mammeri \emph{et al.} \cite{mammeri2014efficient} introduced a method that uses a two-stage approach: utilization of the LBP-Adaboost algorithm in the first stage and an adaptation of HOG-SVM classifier in the second stage for detection of animals. However, there are two major drawbacks of this system: it only considers moose for detection and that too only its side-view. This indicates that other animals that are found on Indian roads such as cows and dogs are not detected by this system. Additionally, this approach can detect only on side-views of the animals and not their front or rear views, thus severely restricting the practical utility of the approach.

\par Ramanan \emph{et al.} \cite{SG_23_3} proposed a different method for animal detection and tracking based on SIFT, which utilizes a texture descriptor that matches it with a predeveloped library of animal textures. The drawback of this approach is that it is limited to footages containing a single animal only and requires the background to be clutter-free. Lastly, Atri Saxena \emph{et al.} \cite{AS_1} has proposed their models based on SSD and faster R-CNN for the detection of animals. The authors have compared the two approaches, however, apart from object detection, no method has been devised that precisely track or pinpoint the presence of the animals on the car's lane. Hence, we determined that an efficient approach was needed which detects a variety of animals found in the Indian roads, overcomes some of the drawbacks in the former approaches, and provides a better detection accuracy for the same.

\section{Methodology}
\label{animal_method}
Stray animals have been a common sighting on Indian roads for a long time. They are commonly sighted on highways, rural roads, and sometimes even in urban cities. These animals are mostly cattle and dogs, but occasionally animals such as elephants and deer can also be observed. Hence, it becomes essential for an autonomous vehicle designed for Indian roads to be able to detect these animals to prevent unwanted accidents and resultant damage to both the vehicle and the animal. The following methodology has been proposed to achieve vehicular collision avoidance with animals based on the input video from a dashboard camera.

\subsection{Animal Detection}
In the first stage, we use the state-of-the-art Mask R-CNN (Region-based Convolutional Neural Networks) model, which is proposed by He \emph{et al.} \cite{DC_37_1}, as the primary object detection model for detecting and identifying animals in the following categories: cat, dog, horse, sheep, cow, elephant, bear, zebra, and giraffe. One of the key reasons for using Mask R-CNN as the object detection model is that it has been trained on the MS COCO dataset \cite{DC_51_1}, which contains 91 common object categories with 82 of them having more than 5,000 labeled instances. This ensures that the aforementioned animal classes have been trained extensively on several thousands of images per class. Since the objective of this support system is to identify animals on Indian roads, we evaluated the ability of the model by primarily testing on frequently found animals such as cows and dogs. 

\par Our Mask R-CNN network has been constructed using a ResNet-101 backbone (bottom-up pathway) through which the input image is passed. Multiple layers are grouped and convolution stages pass through 1$\times$1, 3$\times$3, and 1$\times$1 sized kernels in that order. Each convolution layer is followed by a batch normalization layer and a ReLU activation unit. The layers in the bottom-up pathway are passed through a 1$\times$1 convolution layer so that depth can be downsampled to the corresponding depth of the top-down layer to perform in-place addition. Each feature map is passed through a 3$\times$3 convolution to generate pyramid feature maps. Feature maps are fed to box detection and objectness subnets to generate region proposals. Thereafter, FPN-RoI mapping is performed followed by RoI Align which results in 1024 length vectors for all RoIs from which classification and box regression is performed. The outputs are fed to box regression, classification, and mask branch where regression is performed on each.

\par
\begin{figure}[!h]
	\centering
	\includegraphics[width=0.96\textwidth, height=0.43\textheight]{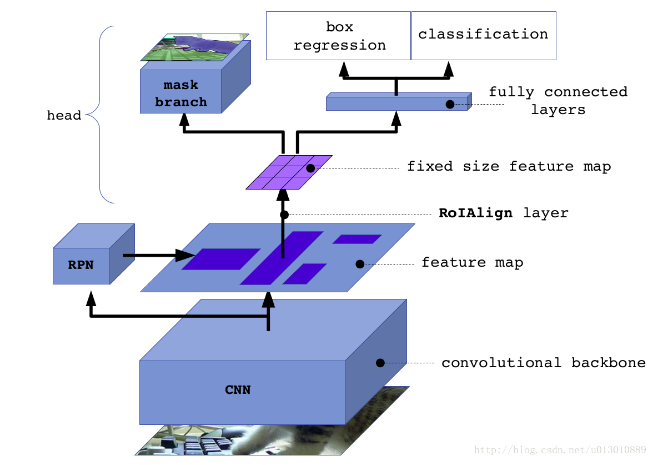}
	\caption{Mask R-CNN architecture \cite{DC_37_mrcnn_fig}}
	\label{fig:mrcnnframework}
\end{figure}

 Additionally, we use Mask R-CNN as it improves Faster R-CNN \cite{ref6} by using Region of Interest (RoI) Align instead of RoI Pooling. The key feature of RoI Align is that it rectifies the location misalignment issue present in RoI Pooling when it takes the input proposals from the Region Proposal Network (RPN) by dividing them into `bins' using bilinear interpolation. Hence, Mask R-CNN improves upon the core accuracy of Faster R-CNN and is therefore used over it. The architecture for Mask R-CNN is illustrated in \figurename~\ref{fig:mrcnnframework}. This stage outputs a dictionary, which contains the class IDs, bounding boxes, detection scores, and the generated masks for a video frame.

\subsection{Lane Detection}
\begin{figure}[!ht]
	\centering
	\includegraphics[width=0.87\linewidth, height=0.675\textheight]{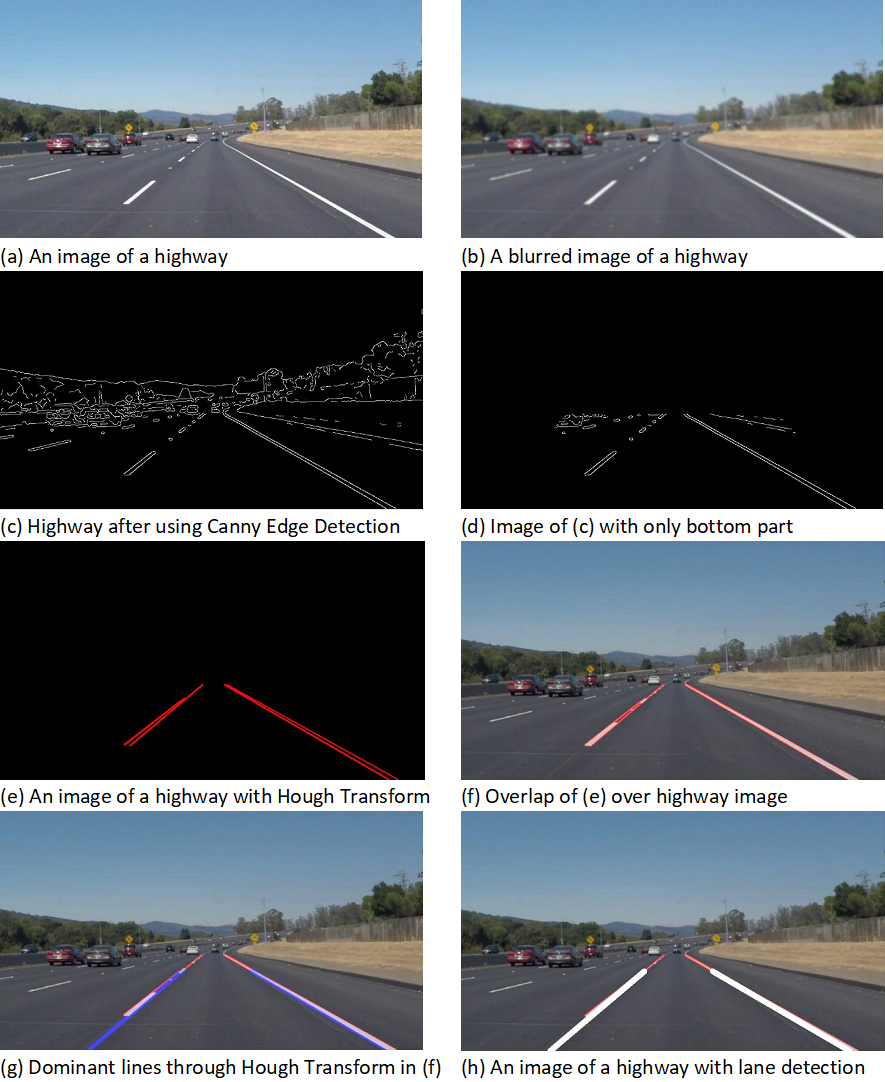}
	\caption{Lane Detection procedure applied on a highway \cite{SG_39_1}}
	\label{fig:lanediagram}
\end{figure}

In this stage, we use an efficient lane detection system, proposed by Aggarwal \cite{SG_39_1}, to identify lane demarcations on the road. This is used to provide feedback to the autonomous vehicle about the animals which are in the path or near to it. For example, in the countryside areas, it is common to see animals in fields on either side of the road in India. These animals do not pose a threat to the vehicle and hence need to be distinguished from those that are of concern, i.e., animals on roads. Hence, we use the lane detection system to deduce whether an animal is in the autonomous vehicle’s lane or not. The location of each animal can either be within the lane of the road or outside it. If it’s within the lane, we notify the vehicle to stop. Otherwise, we perform animal direction analysis and vicinity tracking to provide predictive feedback informing us if an animal outside the lane is about to enter it and possibly collide with the autonomous vehicle. The process of detecting the lane on the highway is shown in \figurename~\ref{fig:lanediagram}.

\subsection{Animal Direction and Vicinity Tracking}
In this stage, we track the location of the detected animals and their direction of movement using a centroid tracking algorithm \cite{SG_43_1}. If an animal is outside a lane but moving towards it, and is in the vicinity of the lane, then we alert the autonomous vehicle even before the animal enters the lane, leading to improved safety. In this step, two criteria need to be met for the animal to be considered as a threat for a possible collision:
\begin{enumerate}[i.]
	\item The animal is outside the detected lanes and is moving towards the lane
	\item The animal is in the vicinity of the vehicle
\end{enumerate}

\subsubsection{Animal Direction Tracking}
After performing object detection, correctly detected animals are retained based on their class IDs and scores by filtering from all detected objects. This paves the way for the next task which is animal tracking over the future time frames of the recording. This is accomplished by utilizing a straightforward yet productive object tracking algorithm called centroid tracking \cite{SG_43_1}. This calculation works by taking the Euclidean distance between the centroids of identified animals over successive frames as clarified further ahead. The centroid tracking algorithm used in this framework can be described as a multiple-step technique which is explained in the following steps:
\begin{enumerate}[i.]
	\item The object centroid for all objects is determined by taking the midpoint of the meeting lines from the middle of the bounding boxes for each distinguished animal. 
	\item The Euclidean distance is determined between the centroids of the newly recognized objects and the current objects. 
	\item The centroid coordinates for the new items are refreshed depending on the least Euclidean distance from the current set of centroids and the newly stored centroid. 
	\item New objects in the field of view are registered by storing their centroid coordinates in a dictionary and appointing fresh IDs. 
	\item Objects which are not noticeable in the current field of view for a predefined set of frames in progression are deregistered from the dictionary.
\end{enumerate}

\par An important presumption of the centroid tracking algorithm is that, though an object will move between the resulting frames of the recording, the separation between the centroid of the same object between successive frames will be less than the separation to the centroid of some other object identified in the frame. The next task in the system is to acquire the trajectories of the tracked animals. These are calculated by finding the contrasts between the centroids of tracked animals for five sequential frames. This is made conceivable by keeping the centroid of every animal in each frame as long as the animal is registered as per the centroid tracking algorithm. The result is a 2D vector, $\boldsymbol{\mu}$, which represents the direction of the animal movement. Next, we determine the magnitude of the vector, $\boldsymbol{\mu}$, as described in Equation \ref{eq:animal_1}.

\begin{equation}\label{eq:animal_1}
\text{magnitude}= \sqrt {\left( {\boldsymbol{\mu}.i } \right)^2 + \left( {\boldsymbol{\mu}.j } \right)^2 }
\end{equation}

\par This vector, $\boldsymbol{\mu}$, is then normalized by dividing it by its magnitude. The vector is stored in a dictionary of normalized direction vectors of every tracked object only if its original magnitude is over a certain threshold. If not, this vector is discarded. This is done to guarantee that minor variations in centroids for static objects do not result in false trajectories.

\subsubsection{Animal Vicinity Tracking}
\label{animal_vicinity}
For the second criterion, we use the horizontal midpoints of the lane and the bounding box of the animal respectively as the comparison parameter to determine whether an animal is in the vicinity of the autonomous vehicle. If the horizontal midpoint of the animal’s bounding box lies between the midpoints of the lane, then we consider that the animal is effectively in the vicinity of the vehicle. The reason is that most of the time, the lane of a road inclines towards the horizon of the road and hence has a gradient which may miss intersecting the bounding box of the animal. This serves as a measure of providing predictive feedback on the possibility of an animal entering the lane of the road.

\subsubsection{Combining Animal Direction and Vicinity Tracking}
\begin{figure}[!ht]
	\centering
	\includegraphics[width=.99\textwidth, height=0.19\textheight]{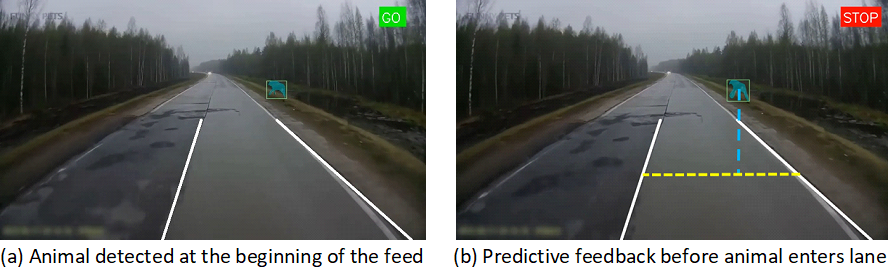}
	\caption{A sequence of steps for predictive feedback of animals outside the lanes}
	\label{fig:ad1}
\end{figure}

Once we track the direction and vicinity of the detected animals, we are able to alert the autonomous vehicle before an animal enters the lane which is moving towards it as shown in \figurename~\ref{fig:ad1}.\par

\subsection{Overall Pipeline}
We construct an overall pipeline by combining the Mask R-CNN model results and lane detection output to get the relevant animal detection alert for the autonomous vehicle. If the animal is within the detected lane, then the support system directly sends a `STOP' alert to the autonomous vehicle. Otherwise, the animals are tracked continuously to determine the direction of their movements. Additionally, the location of the animals' bounding boxes is compared with the midpoints of the lane to check whether the animals are in the vicinity of the path of the autonomous vehicle or not. If an animal is found to be both in the vicinity of the lane and moving towards it, then the support system can send a `STOP' alert to the autonomous vehicle based on a predictive analysis mechanism. Essentially, the objective of this support system is to provide a robust method for preventing potentially unwanted vehicle-to-animal collisions by timely detecting relevant animals on the autonomous vehicle’s path.
\par
\begin{figure}[!ht]
	\centering
	\includegraphics[width=.99\linewidth, height=0.54\textheight]{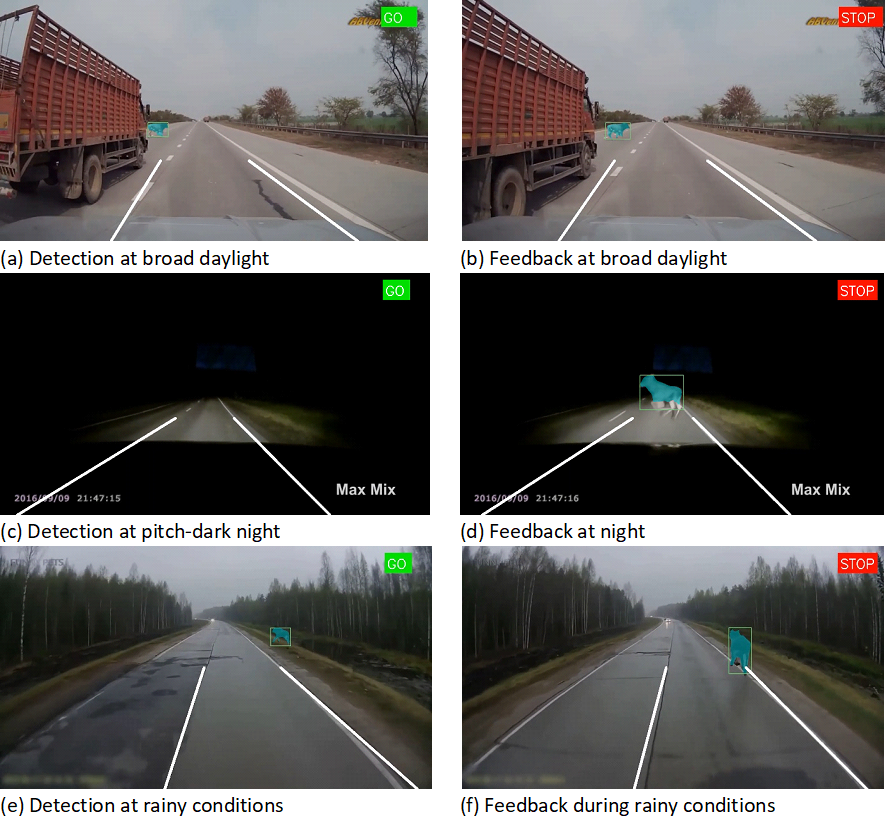}
	\caption{Animal Detection system under various lighting and weather conditions}
	\label{fig:ad2}
\end{figure}

 The proposed framework was tested under various lighting and weather conditions and was observed to perform relatively well for the same as illustrated in \figurename~\ref{fig:ad2}. In this work, we have focused primarily on two most commonly found animals on Indian roads, which are stray cows and stray dogs, for the purpose of evaluation, while at the same time retained the detection of other animals.

\section{Experimental Results}
\subsection{Datasets}
\label{animal_dataset}

\begin{figure}[!ht]
\centering
	\includegraphics[width=0.99\linewidth, height=0.32\textheight]{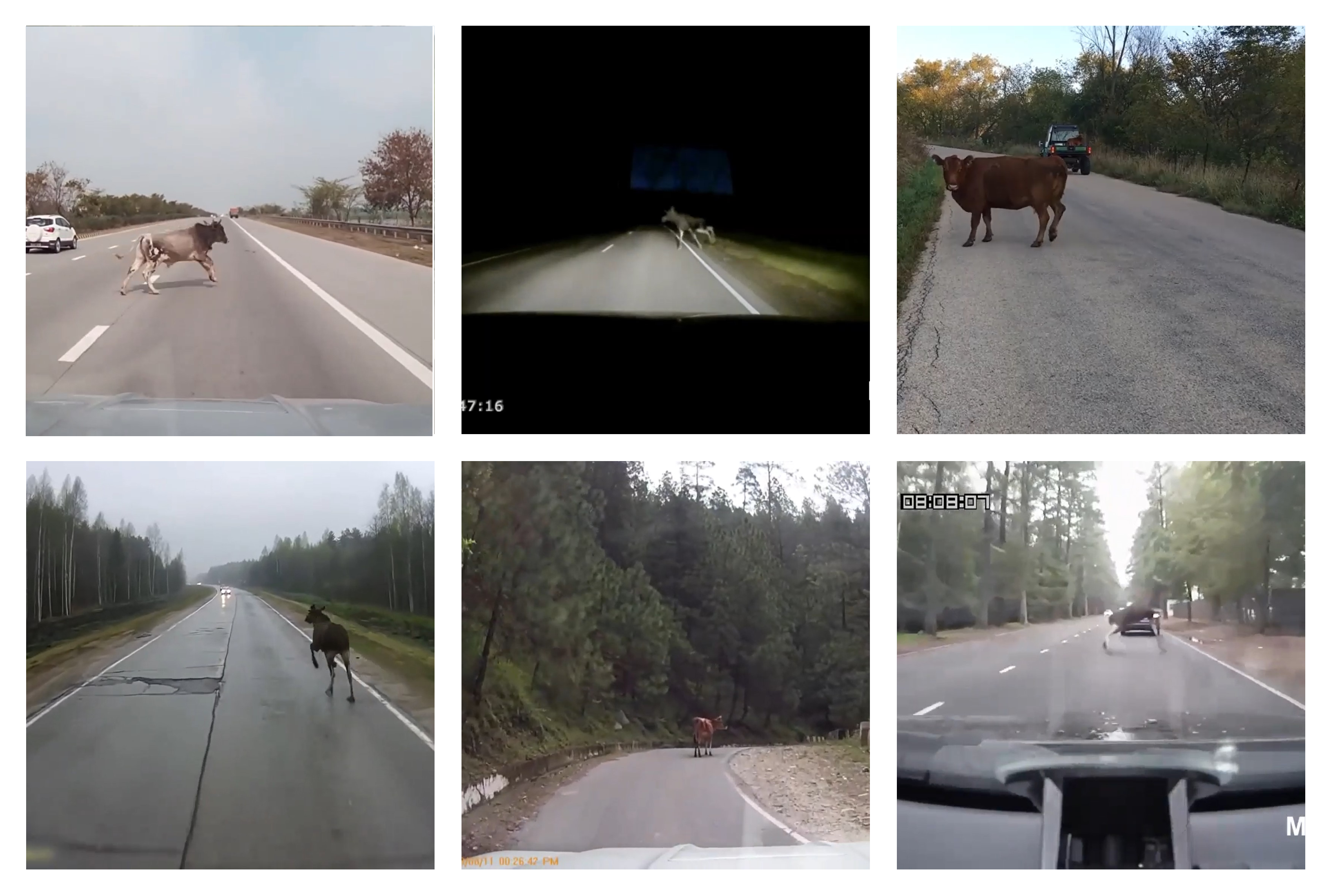}
	\caption{Screenshots taken from test videos of animals sighted on roads dataset}
	\label{fig:animalcollage}
\end{figure}

The following datasets have been used for testing the Vehicle-to-Animal Accident Detection framework:

\begin{enumerate}[i.]
	\item MS COCO Dataset: The Microsoft Common Objects in Context (MS COCO) dataset \cite{DC_51_1} is a well-known dataset that has annotations for instance segmentation and ground truth bounding boxes, which are used to evaluate how well object detection and image segmentation models perform. It contains 91 common objects as classes. Out of the total classes, 82 have 5,000 annotations or more. There are overall 328,000 images with more than 2,500,000 annotated objects. The MS COCO dataset has been used for training the Mask R-CNN framework on multiple object classes including a multitude of animal species.
	
	\item Custom animals sighted on roads dataset: Accuracy of Mask R-CNN and the overall framework has been evaluated on dashcam footage of animals crossing the lanes in front. These videos have been compiled from YouTube. The dashcam videos have been taken at 30 frames per second and consist of footage from various countries around the world, including India, the USA, and Australia. A handful of dataset images are shown in \figurename~\ref{fig:animalcollage}.
\end{enumerate}

\subsection{Results}
Our framework for Vehicle-to-Animal collision avoidance is evaluated in two stages. Firstly, we evaluate how well the Mask R-CNN model identifies the two animal classes on which we based our analysis, i.e. cows and dogs. The evaluation is done using two parameters: precision (Equation \ref{eq:prec}) and recall (Equation \ref{eq:recall}). If the model detects the animals correctly, then we achieve a high recall. If the model distinguishes animals from non-animals, then we get high precision. Generally, precision and recall experience an inverse relationship, where if one increases, the other decreases. The interpolated precision is denoted as $r$, which is computed at each recall level. This is accomplished by selecting the greatest precision measured as shown in Equation \ref{eq:pinterp}. We then generate precision-recall and interpolated precision-recall curves from these values to calculate average precision (Equation \ref{eq:ap}). \par

% \begin{gather}
%     \textit{Precision} = \frac{TP}{TP+FP} \times 100\% \label{eq:prec}\\   
%     \textit{Recall} = \frac{TP}{TP+FN} \times 100\% \label{eq:recall}\\
%     p_{interp}(r) = \max_{r' \geq r} p(r') \quad
%     \label{eq:pinterp}\\
%     \textit{AP} = \sum_{i = 1}^{n - 1} (r_{i + 1} - r_i)p_{interp}(r_{i + 1}) \quad
%     \label{eq:ap}
% \end{gather}

\begin{equation}
\label{eq:prec}
    Precision = TP / (TP+FP)    
\end{equation}
\begin{equation}
\label{eq:recall}
    Recall = TP / (TP+FN)    
\end{equation}
\begin{equation}
\label{eq:pinterp}
    p_{interp}(r) = \max_{r' \geq r} p(r') \quad    
\end{equation}
\begin{equation}
\label{eq:ap}
    AP = \sum_{i = 1}^{n - 1} (r_{i + 1} - r_i)p_{interp}(r_{i + 1}) \quad
\end{equation}

We compute the precision and recall by checking if our model detected the number of animals in the image correctly, as well as the accuracy of the masks. This is done by checking for the degree of the overlap of the predicted bounding boxes (by the Mask R-CNN model) with the actual bounding boxes (of animals). Intersection over Union (IoU) parameter is used to check the extent to which the predicted bounding box corresponds with the actual bounding box. The IoU is calculated by dividing the intersectional area by the area of the union of predicted and actual bounding boxes respectively.
\par
\begin{figure}[!ht]
\vspace{-2em}
\centering
\subfloat[Precision vs Recall Curve]{
\includegraphics[width=.48\textwidth]{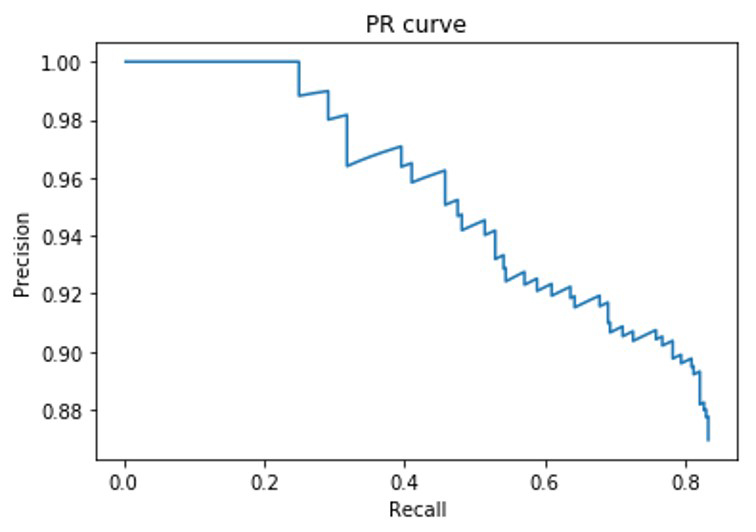}
}
\subfloat[Interpolated Precision vs Recall Curve]{
\includegraphics[width=.48\textwidth]{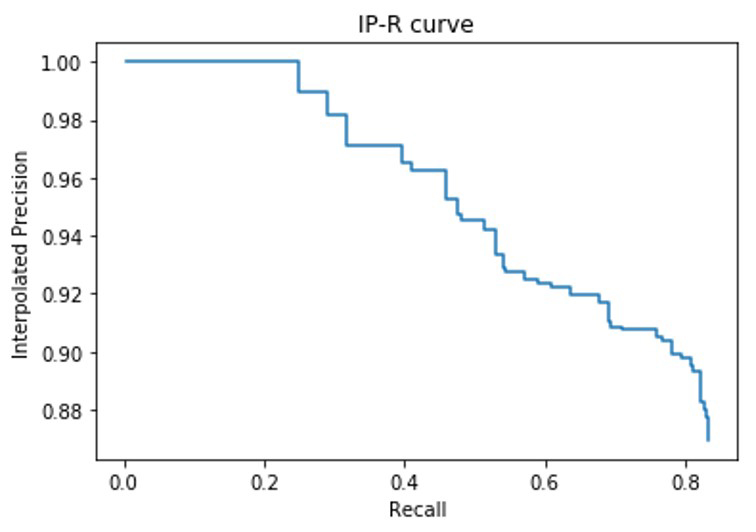}
}
\caption{a) Precision vs Recall curve plotted on the output of Mask R-CNN framework tested on the test set consisting of cows on roads at IoU of 0.5 b) Plotting precision-recall curve after interpolating precision values from the curve a)}
\label{fig:cowresults}
\end{figure}

\begin{figure}[!ht]
\centering
\subfloat[Precision vs Recall Curve]{
\includegraphics[width=.48\textwidth]{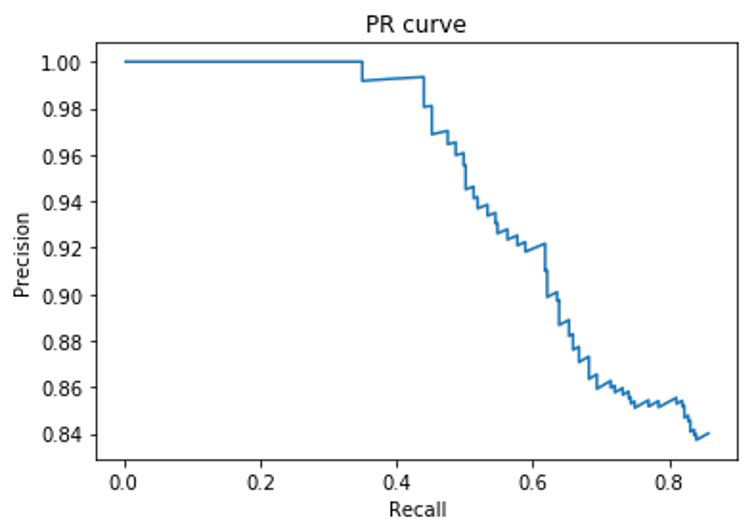}
}
\subfloat[Interpolated Precision vs Recall Curve]{
\includegraphics[width=.48\textwidth]{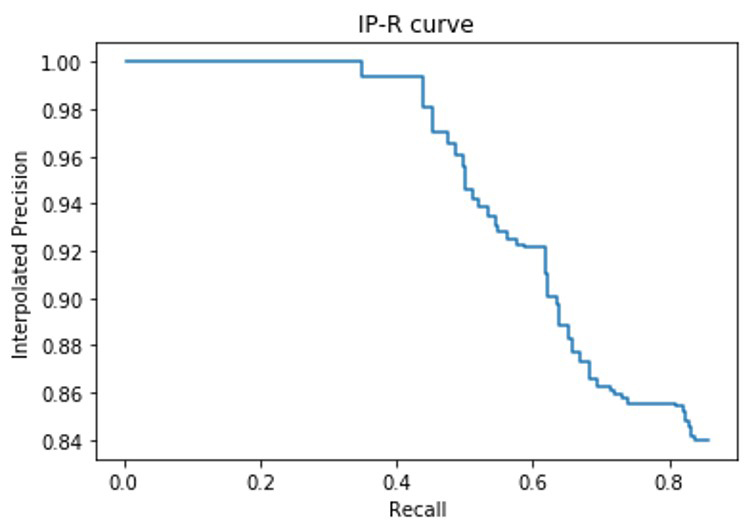} 
}
\caption{a) Precision vs Recall curve plotted on output of Mask R-CNN framework tested on test set consisting of dogs on roads at IoU of 0.5 b) Plotting precision-recall curve after interpolating precision values from curve a)}
\label{fig:dogresults}
\vspace{-2em}
\end{figure}

 We arrange the predicted bounding boxes which match the actual bounding box based on the descending order of their IoU scores. In the case where there isn't a single match, we label this ground truth box as unmatched and move on to evaluate the next set of bounding boxes. This process is repeated. The results were collected for cows and dogs separately.
The results for testing on cows are shown in \figurename~\ref{fig:cowresults} and \tablename~\ref{tb:cowdogval}. It can be seen that the proposed model predicted the cows with a precision of 86.95\% and a recall of 83.33\%. It can also be observed that the model predicts the dogs with a precision of 84.12\% and recall of 85.58\% (\figurename~\ref{fig:dogresults} and \tablename~\ref{tb:cowdogval}).

{\renewcommand{\arraystretch}{1.5}% 
    \vspace{-3em}
	\begin{table}[!ht]
		\centering
		\begin{tabular}{|c|c|c|c|}
		\hline
			Animal & Precision & Recall & Average Precision (AP) \\ \hline
			Cow & 0.8695 & 0.8333 & 79.47\% \\ \hline
			Dog & 0.8412  & 0.8558 & 81.09\% \\\hline
		\end{tabular}
		\caption{\label{tb:cowdogval} Evaluation metrics for Mask R-CNN tested on test set consisting of cows on roads, IOU=0.5}
	\end{table}
	\vspace{-2em}
}

The second stage of the framework deals with predicting whether the detected animal may be on the path of the vehicle based on the predictive feedback mechanism (explained in Section \ref{animal_vicinity}). We evaluate this mechanism based on Potential Accident Detection Ratio (PADR) (Equation \ref{eq:DR_animal}) and False Alarm Rate (FAR) (Equation \ref{eq:FAR_animal}).

\begin{equation}
\label{eq:DR_animal}
\text{PADR} = \frac{\text{Detected potential accident cases}}{\text{Total positive cases in the dataset}}\times 100\%
\end{equation}
\noindent where `Detected potential accident cases' refers to animals near lane which were detected by the predictive feedback mechanism and `Total positive cases in the dataset' refers to the number of instances where animals outside the lane entered the lane eventually.

\begin{equation}
\label{eq:FAR_animal}
\text{FAR} = \frac{\text{Patterns where false alarm occurs}}{\text{Total number of patterns}}\times 100\%
\end{equation}

The results are compared with other studies that have worked in the same domain of detecting animals and have been tabulated as shown in \tablename~\ref{tb:animalresults}.

{\renewcommand{\arraystretch}{1.5}%
    \vspace{-1em}
	\begin{table}[!ht]
		\centering
		\begin{tabular}{|c|c|c|}
			\hline
			Approach & PADR \% & FAR \% \\ \hline
			S. U. {Sharma} \emph{et al.} \cite{SG_23_1} & 80.40 & 0.08 \\ \hline
			Mammeri \emph{et al.} \cite{mammeri2014efficient} & 82.19 & 0.088\\ \hline
			Our Framework & 84.18 & 0.026 \\ \hline
		\end{tabular}
		\caption{Evaluating results of proposed framework in comparison to other frameworks for Potential Vehicle-to-Animal Accident Detection}
		\label{tb:animalresults}
	\end{table}
	\vspace{-2.5em}
} 
Hence, we observed that our framework performs superiorly when evaluated using the Detection Rate and exhibits a lesser False Alarm Rate than similar works. This is achieved by utilizing Mask R-CNN for object detection model and using lane detection to develop a predictive feedback mechanism which helps in detecting animals even before they enter the lanes.

\section{Conclusions and Future Scope}
A novel neural network based framework has been proposed for the avoidance of collision with animals. It uses a multitude of techniques including object detection, lane detection, and object tracking with the help of CNNs and computer vision based techniques to ensure an accurate response. The proposed framework can detect animals that are either in the path of the vehicle or are potentially going to be using a predictive feedback mechanism. It was determined that the system can detect cows with an average precision of 79.47\% and dogs with an average precision of 81.09\%. Moreover, when tested for the correctness of collision detection, our model was able to achieve an accident detection ratio of 84.18\% and a false alarm rate of 0.026\%. Some limitations of this work include are the size of test samples and deviations in lane demarcation detection when the actual lanes haven’t been marked or in cases of blurry/low-resolution videos. These limitations can be addressed in future works.

\bibliographystyle{elsarticle-num}
\bibliography{main}

\begin{thebibliography}{10}
\expandafter\ifx\csname url\endcsname\relax
  \def\url#1{\texttt{#1}}\fi
\expandafter\ifx\csname urlprefix\endcsname\relax\def\urlprefix{URL }\fi
\expandafter\ifx\csname href\endcsname\relax
  \def\href#1#2{#2} \def\path#1{#1}\fi

\bibitem{SG_14_1}
T.~T. Editor,
  \href{https://timesofindia.indiatimes.com/travel/destinations/with-humans-locked-indoors-animals-take-over-the-roads-in-india/as74851938.cms}{With
  humans locked indoors, animals take over the roads in india}, Times of India
  (2020).
\newline\urlprefix\url{https://timesofindia.indiatimes.com/travel/destinations/with-humans-locked-indoors-animals-take-over-the-roads-in-india/as74851938.cms}

\bibitem{SG_14_2}
D.~Bagrecha, A.~K. Rathoure, Biodiversity assessment for asian highway 48 (near
  jaldapara national park) from bhutan to bangladesh passing through india:
  Case study, in: Current State and Future Impacts of Climate Change on
  Biodiversity, IGI Global, 2020, pp. 179--209.

\bibitem{SG_23_1}
S.~U. {Sharma}, D.~J. {Shah}, A practical animal detection and collision
  avoidance system using computer vision technique, IEEE Access 5 (2017)
  347--358.

\bibitem{SG_23_2}
T.~{Burghardt}, J.~{Calic}, Analysing animal behaviour in wildlife videos using
  face detection and tracking, IEE Proceedings - Vision, Image and Signal
  Processing 153~(3) (2006) 305--312.

\bibitem{mammeri2014efficient}
A.~Mammeri, D.~Zhou, A.~Boukerche, M.~Almulla, An efficient animal detection
  system for smart cars using cascaded classifiers, in: 2014 IEEE International
  Conference on Communications (ICC), IEEE, 2014, pp. 1854--1859.

\bibitem{SG_23_3}
D.~Ramanan, D.~A. Forsyth, K.~Barnard, Building models of animals from video,
  IEEE Transactions on Pattern Analysis and Machine Intelligence 28~(8) (2006)
  1319--1334.

\bibitem{AS_1}
A.~Saxena, D.~K. Gupta, S.~Singh, An animal detection and collision avoidance
  system using deep learning, in: G.~S. Hura, A.~K. Singh, L.~Siong~Hoe (Eds.),
  Advances in Communication and Computational Technology, Springer Singapore,
  Singapore, 2021, pp. 1069--1084.

\bibitem{DC_37_1}
K.~{He}, G.~{Gkioxari}, P.~{Dollár}, R.~{Girshick}, Mask r-cnn, in: 2017 IEEE
  International Conference on Computer Vision (ICCV), 2017, pp. 2980--2988.

\bibitem{DC_51_1}
T.-Y. Lin, M.~Maire, S.~Belongie, J.~Hays, P.~Perona, D.~Ramanan,
  P.~Doll{\'a}r, C.~L. Zitnick, Microsoft coco: Common objects in context, in:
  European conference on computer vision, Springer, 2014, pp. 740--755.

\bibitem{ref6}
S.~{Ren}, K.~{He}, R.~{Girshick}, J.~{Sun}, Faster r-cnn: Towards real-time
  object detection with region proposal networks, in IEEE Transactions on
  Pattern Analysis and Machine Intelligence 39~(6) (2017) 1137--1149.

\bibitem{DC_37_mrcnn_fig}
R.~{Jiang},
  \href{https://ronjian.github.io/blog/2018/05/16/Understand-Mask-RCNN}{Understanding-mask
  rcnn}, accessed: 2020-05-30 (2018).
\newline\urlprefix\url{https://ronjian.github.io/blog/2018/05/16/Understand-Mask-RCNN}

\bibitem{SG_39_1}
P.~{Aggarwal}, Detecting lanes with opencv and testing on indian roads, Medium
  (Dec 2016).

\bibitem{SG_43_1}
J.~C. {Nascimento}, A.~J. {Abrantes}, J.~S. {Marques}, An algorithm for
  centroid-based tracking of moving objects, in: Proc. of IEEE International
  Conference on Acoustics, Speech, and Signal Processing (ICASSP), Vol.~6,
  1999, pp. 3305--3308.

\end{thebibliography}
\end{document}